\documentclass{article}
\usepackage[final]{corl_2019} % Uncomment for the camera-ready ``final'' version
\usepackage{amsmath}
\usepackage{amssymb}
\usepackage[ruled,vlined]{algorithm2e}
\usepackage[]{graphicx}
\usepackage{caption}
\usepackage{hyperref}
\usepackage{float}

\title{Weighted Entropy Modification for Soft Actor-Critic}

\author{
  Yizhou~Zhao, Song-Chun Zhu\\
  Department of Statistics\\
  University of California, Los Angeles 
  United States\\ 
  \texttt{yizhouzhao@ucla.edu, sczhu@stat.ucla.edu} \\
}

\begin{document}
\maketitle

%===============================================================================

\begin{abstract}
We generalize the existing principle of the maximum Shannon entropy in reinforcement learning (RL) to weighted entropy by characterizing the state-action pairs with some qualitative weights, which can be connected with prior knowledge, experience replay and evolution process of the policy. We propose an algorithm motivated for self-balancing exploration with the introduced weight function, which leads to state-of-the-art performance on Mujoco tasks despite its simplicity in implementation. 

%show that the weighted entropy can be property embedded into the current maximum entropy reinforcement learning framework and
\end{abstract}


%===============================================================================

\section{Introduction}
Reinforcement learning(RL) studies how an agent engages in its environment and evolves its strategies to get satisfying rewards. Algorithms in reinforcement Learning combined with deep neural networks have proved huge potential in competing Go~\citep{silver2016mastering}, controlling continuous systems for robots~\citep{lillicrap2015continuous}, and playing video games~\citep{mnih2013playing}. However due to the non-determinism in standard benchmark environments combined with variance to the methods, training models and reproducing the results in RL can be extremely hard in tuning hyperparameters, designing network architecture, scaling rewards, and selecting random seeds~\citep{henderson2018deep}.

To improve the performance and robustness of algorithms, changing the learning objective of RL by adding bonus for exploration~\citep{strehl2008analysis, bellemare2016unifying}, regarding the Q-function as a distribution instead of a single value~\citep{tang2018exploration, bellemare2017distributional}, and taking stochastic policies~\citep{sutton2000policy,chou2017improving} all shed light on the use of uncertainty, since RL together with deterministic policy sometimes suffers from overestimation and tends to converge to local optimal solutions without enough exploration.

Among those successful algorithms, maximum entropy RL \citet{haarnoja2017reinforcement,haarnoja2018soft} aim to pursuit the maximum entropy when an agent takes actions, which results in the increment of robustness and the reduction of the expense for sampling. By regularizing entropy~\citep{schulman2015trust,mnih2016asynchronous}, significant process has been made in improving the robustness of RL. Stable model-free algorithms for continuous state and action spaces have been designed under the maximum entropy framework; the pain from tuning the hyperparameters has also been relieved. Besides, the multi-modality can be obtained. %through the policy gradient algorithms by entropy regularization. 

Maximum entropy principle(MEP) emerged for the purpose of proceeding from the laws of physical microscopic mechanism to statistical macroscopic phenomena~\citep{jaynes1957information}. In this article, we cast doubt on the methodology of maximizing Shannon entropy as target of RL even though maximizing Shannon entropy has long been regarded as the routine and recent decades have witnessed a wide range of its successful applications of MEP in statistical models~\citep{wainwright2008graphical}, finance~\citep{zhou2013applications}, computer vision~\citep{zhu1998filters}, natural language processing~\citep{berger1996maximum}, operations research~\citep{abbas2006maximum} and biology~\citep{de2018introduction}. However, since the process of taking actions in reinforcement learning is hardly of total ignorance, there exists great potential if we make less modest guess. Sometimes, auxiliary information such as prior knowledge of the environment, history of sampling or subjective view needs to be considered. We claim the possibility of applying weighted entropy~\citep{guiacsu1971weighted} as the target for RL. Especially, we propose an algorithm that makes better use of auxiliary information by modifying the Shannon entropy term in soft actor-critic as the weighted entropy.

%The primary contribution of this paper is a simple modification for soft actor-critic by considering the weighted entropy under the motivation of denying the previous policy for further exploration. 
The primary contribution of this paper is a simple modification for soft actor-critic by considering the weighted entropy inspired from a more balanced way for exploration. Our experiments
show that our algorithm compares favorably to soft actor-critic, specifically on stochastic control tasks.

\section{Related Work}
This paper extends the work of soft actor-critic~\citep{haarnoja2018soft,Haarnoja:EECS-2018-176}, which is based on the maximum entropy (inverse) reinforcement learning framework~\citep{haarnoja2017reinforcement, o2016combining, schulman2017equivalence, ziebart2008maximum} and the off-policy learning~\citep{nachum2017bridging, degris2012off}. Existing soft actor-critic~\citep{haarnoja2018soft} reaches the competitive and stable performance on model-free reinforcement learning. It shares the the actor-critic framework with the commonly utilized RL algorithms such as PPO~\citep{schulman2017proximal} and ACKTR~\citep{wu2017scalable}. We focus our work on continuous control with infinite horizon as well. The common algorithms in such settings include DDPG~\citep{lillicrap2015continuous} and TD3~\citep{fujimoto2018addressing}.

Recently, achieving maximum general entropy reinforcement learning has been proposed by \citet{lee2019tsallis} and \citet{chen2019off}. The idea is to generalize the entropy measure i.e. Shannon entropy into Tsallis entropy and R\'{e}nyi entropy. Even though theoretical analysis and experiment result show that incorporating Tsallis or R\'{e}nyi entropy can be really effective than in some different experimental settings, the motivation behind generalizing Shannon entropy into non-Shannon ones remains unclear. In our opinion, the objective of achieving maximum entropy in RL is to learn a nearly optimal solution which preserves multi-modality. %We name it as \textbf{utility learning}. 
Our work is motivated by weighted entropy(WE)~\citep{guiacsu1971weighted, suhov2016basic}, which takes into account values of different outcomes, i.e., makes entropy context-dependent, through the weight function~\citep{kelbert2017weighted}. \citet{zhao2019maximum} proposed the similar idea of maximizing the entropy weighted by rewards, whereas we aim to design more arbitrary weight function and more flexible algorithms.  

\section{Preliminaries}
\subsection{Maximum entropy reinforcement learning and soft actor critic}
We consider the Markov decision process(MDP), defined as a tuple $(\mathcal{S}, \mathcal{A}, p, r)$. $\mathcal{S}$ is the state space  and $\mathcal{A}$ the action space. Transition function $p$ is defined on $\mathcal{S}\times \mathcal{S}\times \mathcal{A}$ represents the probability(density) of transiting to next state $s_{t+1} \in \mathcal{S}$ given the current state $s_t \in \mathcal{S}$ and action $a_t \in \mathcal{A}$. $r$ is the reward function $\mathcal{S} \times \mathcal{A} \to \mathbb{R}$. We also denote $\rho^\pi(s_t)$ and $\rho^\pi(s_t, a_t)$ as the expected state and state-action visiting counts along trajectories induced by policy $\pi$. Then, the MDP problem can be formulated as finding the policy $\pi$ defined on $\mathcal{S} \times \mathcal{A}$ maximizes
\begin{align}
J(\pi) = \mathbb{E}_{(s_t, a_t)\sim \rho^\pi}[\sum_{t = 0}^T\gamma^tr(s_t, a_t)]
\end{align}
where $\gamma$ is a discount factor and $T$ is the maximum time frame. Standard RL dives directly into maximizing the rewards without sufficient exploration and such objective cannot catch multiple modes for near optimal solutions. Maximum entropy reinforcement learning~\citep{haarnoja2017reinforcement} favors stochastic policies by augmenting the objective with entropy:
\begin{equation}
J(\pi)=\sum_{t=0}^{T} \mathbb{E}_{\left(s_{t}, a_{t}\right) \sim \rho^{\pi}}\left[\gamma^t(r\left(s_{t}, a_{t}\right)+\alpha \mathcal{H}\left(\pi\left(\cdot | s_{t}\right)\right))\right]
\end{equation}
The temperature parameter $\alpha$ determines the trade-off between maximizing rewards and Shannon entropy $\mathcal{H}$. %When $\alpha \to 0$, the standard objective of RL recovers and when $\alpha \to \infty$, the agent tends to take actions uniformly at each state. %In practice, the desired temperature is obtained by solving the constrained optimization problem:
% \begin{align}
% %     \max_\pi J(\pi) \text{ s.t. }\mathbb{E}_{\left(\mathbf{s}_{t}, \mathbf{a}_{t}\right) \sim \rho_{\pi}}\left[-\log \left(\pi_{t}\left(\mathbf{s}_{t} | \mathbf{s}_{t}\right)\right)\right] \geq \overline{\mathcal{H}}
% % \end{align}
% % $\overline{H}$ is the desired minimum expected entropy. From an approximate dynamic approach and considering $\alpha_t$ as a dual variable, we can solve the temperature $\alpha_t^*$ after solving for optimal policy $\pi_t^*$(see details in \cite{Haarnoja:EECS-2018-176}):
% % \begin{equation}
% % \alpha_{t}^{*}=\arg \min _{\alpha_{t}} \mathbb{E}_{\mathbf{a}_{t} \sim \pi_{t}^{*}}\left[-\alpha_{t} \log \pi_{t}^{*}\left(\mathbf{a}_{t} | \mathbf{s}_{t} ; \alpha_{t}\right)-\alpha_{t} \overline{\mathcal{H}}\right]
% % \end{equation}
Soft actor-critic~\citep{haarnoja2018soft} solves the maximum entropy objective from three parts: soft Q-function $Q(s_t, a_t)$, value function $V(s_t)$ and policy $\pi(a_t|s_t)$. Starting at any function $Q: \mathcal{S}~\times~\mathcal{A}~\to~\mathbb{R}$ and the modified Bellman backup operator $\mathcal{T}^\pi$ with respect to policy $\pi$ given by
\begin{equation}
\mathcal{T}^{\pi} \circ Q\left(s_{t}, a_{t}\right) := r\left(s_{t}, a_{t}\right)+\gamma \mathbb{E}_{s_{t+1} \sim p}\left[V\left(s_{t+1}\right)\right]
\end{equation}
where
\begin{equation}
V\left(\mathbf{s}_{t}\right)=\mathbb{E}_{\mathbf{a}_{t} \sim \pi}\left[Q\left(s_{t}, a_{t}\right)-\alpha\log \pi\left(a_{t} | s_{t}\right)\right]
\end{equation}
Under the common off-policy settings~\citep{silver2014deterministic, degris2012off}, suppose the trajectories are sampled from another policy $\tilde{\pi}(a_t|s_t) \neq \pi(a_t|s_t)$, the target of soft actor-critic is to maximize 
% % \begin{equation}
% % \begin{aligned} \max_{\pi} J_{\tilde{\pi}}\left(\pi\right) =\int_{\mathcal{S}} \rho^{\tilde{\pi}}(s) V^{\pi}(s) \mathrm{d} s =\int_{\mathcal{S}} \int_{\mathcal{A}} \rho^{\tilde{\pi}}(s) \pi(a | s) Q^{\pi}(s, a) \mathrm{d} a \mathrm{d} s \end{aligned}
% % \end{equation}
% % Instead, soft actor-critic maximizes the performance of the target policy averaged on the state-action pairs, i.e.,
\begin{equation}
\begin{aligned} %\max_{\pi}
J_{\tilde{\pi}}\left(\pi\right) = \int_{\mathcal{S}} \int_{\mathcal{A}} \rho^{\tilde{\pi}}(s, a) Q^{\pi}(s, a) \mathrm{d} a \mathrm{d} s %=\int_{\mathcal{S}} \int_{\mathcal{A}} \rho^{\tilde{\pi}}(s) \tilde{\pi}(a | s) Q^{\pi}(s, a) \mathrm{d} a \mathrm{d} s 
\end{aligned}
\end{equation}
where $Q^\pi$ is the soft Q-function converged from operation (3). Considering the updating rules and mean square error, value function $V(s_t)$ and soft Q-function $Q(s_t, a_t)$ can be updated. Instead of calculating the gradient to maximize the performance function $J_{\tilde{\pi}}\left(\pi\right)$, soft actor-critic maximizes the target $J_{\tilde{\pi}}\left(\pi_{\theta}\right)$ by minimizing the KL-divergence between current policy and a Boltzmann distribution with energy term $Q(s_t, \cdot)/\alpha$:
\begin{equation}\label{eq:KL_rule}
\pi_{\mathrm{new}}=\arg \min _{\pi^{\prime}} \mathrm{D}_{\mathrm{KL}}\left(\pi^{\prime}\left(\cdot | s_{t}\right) \bigg\| \frac{\exp \left(\frac{1}{\alpha}Q^{\pi_{\mathrm{old}}}\left(s_{t}, \cdot\right)\right)}{Z^{\pi_{\mathrm{old}}}\left(s_{t}\right) )}\right)
\end{equation}
$Z^{\pi_{\mathrm{old}}}\left(\mathbf{s}_{t}\right)$ is the partition function. \citet{haarnoja2018soft} proves that the above updating rule guarantees the improvement of Q-function: $Q^{\pi_{\mathrm{ new}}}\left(s_{t}, a_{t}\right) \geq Q^{\pi_{\mathrm{old}}}\left(s_{t}, a_{t}\right)$, thus $J_{\tilde{\pi}}\left(\pi_\mathrm{new}\right) \geq J_{\tilde{\pi}}\left(\pi_\mathrm{old}\right)$. 

\subsection{From Shannon entropy to weighted entropy}
Entropy, in physics and information theory, is the measure for randomness or dispersion. Usually, \textit{entropy} is regarded as Shannon entropy,
\begin{equation}
H(X)=-\sum_{i=1}^{n} p_{i} \log p_{i}
\end{equation}
which is characterized by the so called Shannon-Khinchin axioms(see details in Appendix A and~\citep{khinchin2013mathematical}). Shannon entropy is unbiased and context-free, i.e., does not depend on the nature of the outcomes, but only on the probabilities $p$~\citep{suhov2016basic}.  Even though the Shannon entropy is widely accepted and shares a lot of successful applications in reinforcement learning. In many situations, the context-free setting seems to be insufficient. Imagine that we prefer an agent to explore more in an environment. Then, the actions that reaches new states are more favorable than the actions that transit the agent to the states which are visited before. And in my cases, we may find an agent taking useless actions based on human knowledge and we need a way to make those cases less favorable. Motivated by such consideration of making better use of auxiliary information, we start our study in weighted entropy, making the state-action pairs context-dependent.

Weighted entropy~\citep{guiacsu1971weighted} is the measure of information supplied by a probabilistic experiment whose elementary events are characterized both by their objective probabilities and by some qualitative (objective or subjective) weights. Weighted entropy evaluates the amount of information supplied by a probability space from the probabilities of the events and objective or subjective weights from the experimenters. 

The weighted entropy is defined as(see Appendix A and~\citep{guiacsu1971weighted})
\begin{align}
    \mathcal{H}^w(p) = H_n(w_1,...,w_n; p_1,...,p_n) = -\sum_{k=1}^nw_kp_k\log p_k
\end{align}
where $w = (w_1,...,w_n),p=(p_1,...,p_n)$. %In continuous cases, we consider the generalization $\mathcal{H}^w(p) = \int w\cdot p\log p~dp$.
\citet{guiacsu1971weighted} proves that weighted entropy defined on a finite set gets the maximum $\zeta+\sum_{i=1}^{n} w_{i} e^{-\left(\zeta / w_{i}\right)-1}$  if and only if $
p_{i}=e^{-\left(x / w_{i}\right)-1}; i=1, \ldots, n$ where $\zeta$ is the solution of the equation 
$\sum_{i=1}^{n} e^{-\left(\zeta / w_{i}\right)-1}=1$. 

\section{Weighted entropy modification for soft actor-critic architecture}
We will follow the derivation of soft actor-critic from the policy iteration, which starts from a weighted entropy augment for the Q-function as well as the value function. In the beginning, we will set the weights to be constant in the general case and then present different designs for the weights and illustrate the feasibility based on theoretical analysis.

\subsection{Policy iteration}
Our goal is to find the optimal policy that maximizes reward with the weighted entropy,
\begin{align}
J^w(\pi)&=\sum_{t=0}^{T} \mathbb{E}_{\left(s_{t}, a_{t}\right) \sim \rho^{\pi}}\left[\gamma^t(r\left(s_{t}, a_{t}\right)+\alpha \mathcal{H}^w\left(\pi\left(\cdot | s_{t}\right)\right))\right] 
\end{align}
Then weighted entropy is defined as 
\begin{align}
\mathcal{H}^w(\pi(\cdot | s) = -\int w(s,a)\pi(a|s)\log \pi(a|s) da    
\end{align}
The weight $w$ is a function $\mathcal{S} \times \mathcal{A} \to \mathbb{R}^+$. In this section, we discuss the situation that $w$ remains unchanged during the policy updating process. Dynamic and arbitrary design of weights will be discussed in the next section. Similar to soft actor-critic, initialized from any function $\mathcal{S} \times \mathcal{A} \to \mathbb{R}$, soft Q-function is repeatedly updated under the current policy
\begin{align}
    \tilde{\mathcal{T}}^\pi \circ Q(s_t, a_t) &= r(s_t, a_t) + \gamma \mathbb{E}_{s_{t+1} \sim p}[V(s_{t+1})] \\
    \text{where }V(s_t) &= \mathbb{E}_{a_t\sim \pi}[Q(s_t, a_t) - \alpha w(s_t, a_t)\log \pi(a_t|s_t)]
\end{align}
Consider the updating operation $\tilde{\mathcal{T}}^\pi$ in equation (11) under the condition that we do not make any change with respect to~$w$. Then the sequence $\{Q^0, Q^1, ..., Q^n,...\}$, where $Q^{k+1} = \tilde{\mathcal{T}}^\pi \circ Q^k = (\tilde{\mathcal{T}}^\pi)^{k+1} \circ Q^0$, converges to the soft Q-value $Q^*$ of $\pi$ under the weights $w$. The proof exactly follows \citet{haarnoja2018soft}.\hfill $\square$

In the policy improvement step, the soft actor-critic tires to minimize the KL divergence between the new policy and the energy-base policy according to rule \eqref{eq:KL_rule}. %In practice, rather than to find the exact minimum, soft actor-critic only requires the updated new policy is closer to the optimal,
%\begin{align} \mathrm{D}_{\mathrm{KL}}\left(\pi^{new}\left(\cdot | \mathbf{s}_{t}\right) \bigg\| \frac{\exp \left(Q^{\pi_{\mathrm{old}}}\left(\mathbf{s}_{t}, \cdot\right)\right)}{Z^{\pi_{\mathrm{old}}}\left(\mathbf{s}_{t}\right) )}\right) \leq \mathrm{D}_{\mathrm{KL}}\left(\pi^{old}\left(\cdot | \mathbf{s}_{t}\right) \bigg\| \frac{\exp \left(Q^{\pi_{\mathrm{old}}}\left(\mathbf{s}_{t}, \cdot\right)\right)}{Z^{\pi_{\mathrm{old}}}\left(\mathbf{s}_{t}\right) )}\right)
%\end{align}
%and the improvement of soft Q-function is guaranteed, i.e., $Q^{\pi_{new}}(s_t, a_t) \geq Q^{\pi_{old}}(s_t, a_t)$ for any $(s_t, a_t)\in \mathcal{S} \times \mathcal{A}$ with $\mathcal{A} \leq \infty$. 
One possible generalization is to consider the weighted Kullback-Leibler divergence~\citep{suhov2016basic},
\begin{align}\label{eq:weighted-KL}
    \mathrm{D}_{\mathrm{KL}}^w(p||q) = \int w(x)p(x)\log \frac{p(x)}{q(x)} dx
\end{align}
the following lemma suggests that the policy iteration needs to satisfy an extra constraint to get monotone increment for Q-function.

\textbf{Lemma 1}: Let $\pi_{\mathrm{old}}$ be the policy to be updated and if $\pi_{\mathrm{new}}$ satisfies conditions
\begin{align}\label{eq:weight-updated}
    \mathrm{D}_{\mathrm{KL}}^w\left(\pi_{\mathrm{new}}\left(\cdot | s_{t}\right) \bigg\| \frac{\exp \left(\frac{Q^{\pi_{\mathrm{\mathrm{old}}}}\left(s_{t}, \cdot\right)}{\alpha w(s_t,\cdot)}\right )}{Z^{\pi_{\mathrm{\mathrm{old}}}}\left(s_{t}\right)}\right) &\leq \mathrm{D}_{\mathrm{KL}}^w\left(\pi_{\mathrm{old}}\left(\cdot | s_{t}\right) \bigg\| \frac{\exp \left(\frac{Q^{\pi_{\mathrm{\mathrm{old}}}}\left(s_{t}, \cdot\right)}{\alpha w(s_t,\cdot)}\right )}{Z^{\pi_{\mathrm{\mathrm{old}}}}\left(s_{t}\right) }\right)
\end{align}
\begin{align}\label{eq:constraint1}
    \text{and }\int w(s_t,a_t)\pi_{\mathrm{new}}(a_t | s_t)da_t &=  \int w(s_t,a_t)\pi_{\mathrm{old}}( a_t | s_t)da_t
\end{align}
Then $Q^{\pi_{\mathrm{new}}}(s_t, a_t) \geq Q^{\pi_{\mathrm{old}}}(s_t, a_t)$ for any $(s_t, a_t)\in \mathcal{S} \times \mathcal{A}$ with $\mathcal{A} \leq \infty$.

\textit{Proof}. See Appendix B.\hfill$\square$

Notice that $Z^{\pi_{\mathrm{old}}}(s_t)$ is the partition function and does not contribute to the gradient with respect to new policy. However, because constraint \eqref{eq:constraint1} is intractable during optimization steps, one practical approximation is to consider the updating rule \eqref{eq:weight-updated} only,which may introduce bias and unpredictable behaviors.
% \begin{align}
%     \min_\pi   & ~\mathrm{D}_{\mathrm{KL}}^w\left(\pi\left(\cdot | \mathbf{s}_{t}\right) \bigg\| \frac{\exp \left(Q^{\pi_{\mathrm{old}}}\left(\mathbf{s}_{t}, \cdot\right)\right/w(s_t, \cdot))}{Z^{\pi_{\mathrm{old}}}\left(\mathbf{s}_{t}\right) )}\right) \\
%     \text{s.t. } &\mathbb{E}_{a_t \sim \pi}[w(s_t, a_t)] = \bar{C}
% \end{align}
% where $\bar{C}$ is a constant. 
% The optimization problem with the constraint is usually intractable during policy iteration. one approximation is to ignore the constrain; 

In this article, we follow a better updating rule proposed by \citet{chen2019off}, i.e.
\begin{equation} \label{eq:policy1}
\pi_\mathrm{new} = \max _{\pi^{\prime}} \mathbb{E}_{a \sim \pi^{\prime}(\cdot | s_t)}\left[Q^{\pi_\mathrm{old}}(s_t, a)+\alpha \mathcal{H}^w\left(\pi^{\prime}(\cdot | s_t)\right)\right]
 \end {equation}

\textbf{Lemma 2}: Let $\pi_{\mathrm{old}}$ be the policy to be updated and if $\pi_{\mathrm{new}}$ satisfies the condition
\begin{align}\label{eq:renew}
    \mathbb{E}_{a \sim \pi_{\mathrm{new}}}\left[Q^{\pi_\mathrm{old}}(s_t, a)+\alpha \mathcal{H}^w\left(\pi_{\mathrm{new}}(\cdot | s_t)\right)\right] \geq \mathbb{E}_{a \sim \pi_{\mathrm{old}}}\left[Q^{\pi_\mathrm{old}}(s_t, a)+\alpha \mathcal{H}^w\left(\pi_{\mathrm{old}}(\cdot | s_t)\right)\right]  
\end{align}
Then $Q^{\pi_{\mathrm{new}}}(s_t, a_t) \geq Q^{\pi_{\mathrm{old}}}(s_t, a_t)$ for any $(s_t, a_t)\in \mathcal{S} \times \mathcal{A}$ with $\mathcal{A} \leq \infty$.

\textit{Proof}. See Appendix B.\hfill$\square$

In the general actor-critic settings, the updating rule for Q function relies on current policy $\pi$, and the updating rule for policy relies on current Q function. Suppose the general updating rules as follow,
\begin{align}
    Q_{\mathrm{new}} &= \mathcal{T}^\pi \circ Q_{\mathrm{old}} \\
    \pi_{\mathrm{new}} &= \Phi^Q \circ \pi_{\mathrm{old}}
\end{align}
\citet{haarnoja2018soft} argue that if $\mathcal{T}^\pi$ is a \textbf{contraction}(i.e. $\lim_{n\to\infty} (\Gamma^\pi)^n \circ Q = Q^{\pi^*}$) and $\Phi^Q$ makes \textbf{improvement}(i.e. $Q^{\pi^*_{new}} \geq Q^{\pi^*_{old}}$), learning objective $J(\pi)$ converges to its optimal. Similarly, $J^w(\pi)$ converges to it optimal with weight function $w$.

\subsection{Dynamic weight function}
So far in the derivation the weight function $w$ remains unchanged in the training process. However, in many scenarios, the weight function needs adjustment. For example, in discrete state and action space $\mathcal{S} \times \mathcal{A}$, the weight of a certain state-action pair $(s, a)$ can be relatively large in the beginning and gradually becomes smaller because of the losing interest for exploration.

We focus our discussion about on the large-horizon or infinite-horizon Markov decision process. First, we assume that arbitrary designs of weights which are bounded within interval $[0, 1]$. Thus, the weighted entropy $\mathcal{H}^w$ is bounded. Suppose currently the largest number of time steps the agent can survive under under its policy is $T_0$. The difference between the learning objectives $J^{w_1}$ and $J^{w_2}$ is bounded when the weight function changes from $w_1$ to $w_2$.
\begin{align}
    \bigg|\sum_{t=0}^{T_0} \mathbb{E}_{\left(s_{t}, a_{t}\right) \sim \rho^{\pi}}\big[\gamma^t\alpha \big(\mathcal{H}^{w_1}\left(\pi\left(\cdot | s_{t}\right)\right)-\mathcal{H}^{w_2}\left(\pi\left(\cdot | s_{t}\right)\right)\big)\big]\bigg|\large < C_0
\end{align}
If the policy iteration could make the agent survive longer to $T_1$ and get a new policy $\pi_{\mathrm{new}}$ with the improvement greater than constant $C_0$,
\begin{align}
    \sum_{t=T_0+1}^{T_1} \mathbb{E}_{\left(s_{t}, a_{t}\right) \sim \rho^{\pi_{\mathrm{new}}}}\Big[\gamma^t(r(s_t, a_t) + \alpha \mathcal{H}^{w_1}\left(\pi_{\mathrm{new}}\left(\cdot | s_{t}\right)\right))\Big]\large > C_0
\end{align}

The local improvement of the policy can still obtain a better policy because the total return is dominated by reward $r$ rather than entropy $\mathcal{H}^w$ in the long term. Applying the dynamic weight function brings more flexibility compared with the soft actor-critic.

\subsection{Algorithm}                 The weight function introduced brings both uncertainty and flexibility, which is the most challenge part. The full algorithm alternates among policy iteration, Q-function evaluation and weight function adjustment.  

Let $\mathcal{D}$ be the distribution of historical state-action pairs, which usually are stored in replay buffer. Initialize the parameters $\theta$ and $\phi$ for Q-function $Q_{\theta}$ and policy $\pi_{\phi}$ as well as weight function $w$. The parameters of Q-function are trained to minimize mean square loss:
\begin{equation}
    L_Q(\theta) = \mathbb{E}_{(s_t, a_t)\sim \mathcal{D}}\big[(Q_\theta(s_t, a_t)-\hat{Q}_\theta(s_t, a_t))^2\big]
\end{equation}
with 
\begin{equation}
    \hat{Q}_\theta(s_t, a_t) = r(s_t, a_t) + \gamma [Q_\theta(s_{t+1}, a_{t+1}) - \alpha \cdot w(s_{t+1}, a_{t+1})\log \pi_\phi(a_{t+1} | s_{t+1})]
\end{equation}
where $(s_t, a_t, r(s_t, a_t), s_{t+1})$ is sampled from $\mathcal{D}$ and $a_{t+1}$ is sampled from policy $\pi_{\phi}(\cdot|s_{t+1})$. Alternatively, policy iteration follows rule \eqref{eq:policy1}:
\begin{equation}\label{eq:policy}
    L_\pi(\phi) = -\mathbb{E}_{a_t \sim \pi_\phi}\big[Q_\theta(s_t, a_t) + \alpha \mathcal{H}^w(\pi_\phi(\cdot | s_t))\big]
\end{equation}

The weight function $w$ distinguishes our algorithm from standard soft actor-critic, which can be related to human knowledge, replay buffer or policy. Our modification is applied to Soft Actor-Critic(SAC) to form Weighted Entropy Soft Actor-Critic(WESAC). In stead of parameterizing weight function $w$ to make it trainable, we derives it directly from historical policy under the motivation of encouraging further exploration.  Let $\pi_{\mathrm{delay}}$ record the previous policy, i.e., the policy which was $k_{\mathrm{delay}}$ epochs ago. We define \textbf{self-balancing  weight function} for as
\begin{equation}\label{eq:weight}
    w(s_t, a_t) = 1 -  \frac{\pi_{\mathrm{delay}}(a_t|s_t)}{\max_a\pi_{\mathrm{delay}}(a|s_t)}
\end{equation}
where $a_t$ is sampled from current policy $\pi_\phi$. Intuitively, the above weight function makes balance between the current policy $\pi_\phi$ and previous policy $\pi_{\mathrm{delay}}$ like this: it gives relatively larger weight to the state-action pair which is unlikely to be sampled from previous policy and less weight to the pair which has a large likelihood to be sampled previously. In our opinion, weighted entropy soft actor-critic(WESAC) which applies the self-balancing weight function tends to take more meaningful exploration, since it is linked to historical information(historical policy) and gives relatively larger bonus to the action which is less likely to be taken previously.

In practice, $\pi_{\mathrm{delay}}$ is approximated by the $\pi_{\bar{\phi}}$, where $\bar{\phi}$ is the moving average of $\phi$.

\begin{algorithm}[H]
\SetAlgoLined
%\KwResult{}
 Initialize parameters $\theta$ for Q-function $Q_\theta$, $\phi$ for policy $\pi_\phi$, smoothing coefficient $\eta$ for $\bar{\phi}$;\\
 \For{each iteration}{
    \For{each environment step}{
        $a_t \sim \pi_\phi(a_t | s_t)$ \\
        $s_{t+1} \sim p(s_{t+1}|s_t, a_t)$ \\
        $\mathcal{D} \leftarrow \mathcal{D}\cup \{(s_t, a_t, r(s_t, a_t), s_{t+1})\}$
    }
  \For{each gradient step}{
        $\theta \leftarrow \theta-\lambda_{Q} \hat{\nabla}_{\theta} L_{Q}\left(\theta\right)$\\
        $\phi \leftarrow \phi-\lambda_{\pi} \hat{\nabla}_{\phi} L_{\pi}(\phi)$ \\
        $\bar{\phi} = \eta\phi + (1-\eta)\bar{\phi}$\\
         update weight function $w$ by formula \eqref{eq:weight} \\
        %$\alpha \leftarrow \alpha \lambda_\alpha\hat{\nabla}_{\alpha} L_\alpha$ If \textbf{temperature-tuning}\\
  }
 %for every $t_{w}$ steps\\
%   \eIf{condition}{
%   instructions1\;
%   instructions2\;
%   }{
%   instructions3\;
%   }
 }
 
 \caption{Weighted Entropy Soft Actor-Critic}
\end{algorithm}
Besides, the double Q-learning trick~\citep{fujimoto2018addressing, van2016deep} and the reparameterization trick~\citep{silver2014deterministic} are applied for training tasks in continuous control.

\section{Experimental Results}
To examine the performance our algorithm, we run on the MuJoCo~\citep{todorov2012mujoco}(v2) stochastic control tasks with  provided by OpenAI Gym~\citep{brockman2016openai}. We compare our methods to the state-of-the-art soft actor-critic and run our algorithms on the same platform\footnote{MuJoCo experiments of Version two: https://
github.com/vitchyr/rlkit}. Our algorithms share exactly the same hyper-parameters with SAC except the weight function. 

In recent years, many works utilized other physics engines to simulate system dynamics of control tasks, such as PyBullet~\citep{tan2018sim}. However, different settings of network architectures, reward scales, random seeds and trials in the new environments make it hard to judge the improvement compared with the existing methods.

\begin{table}[h]
\centering
\begin{tabular}{|c|c|c|c|c|c|}
\hline
Hopper & Humanoid & HalfCheeta & Walker2d & Ant    & HumanoidStandup \\ \hline
5.36\% & 0.00\%   & 22.48\%    & 7.41\%   & 6.65\% & 14.76\%         \\ \hline
\end{tabular}
\vspace{0.5em}
\caption{Average improvement of WESAC compared with SAC on MuJoCo tasks. The results are based on the average exploration returns after training for two million steps. Each experiment is repeated for five times.}
\end{table}
\vspace{-1.5em}
Our algorithm is easy to implement and the only extra term is the weight function. Six canonical tasks involved in our experiments include Hopper, Ant, Humanoid, HalfCheeta, Walker2d and HumanoidStandup. Task Swimmer(v2) is not taken into account because of the mechanism of random reward makes both SAC and WESAC hard to select random seeds to obtain satisfying results.

\begin{figure}[h!tbp]
\begin{minipage}{0.499\textwidth}
\includegraphics[height=42mm, width=64mm]{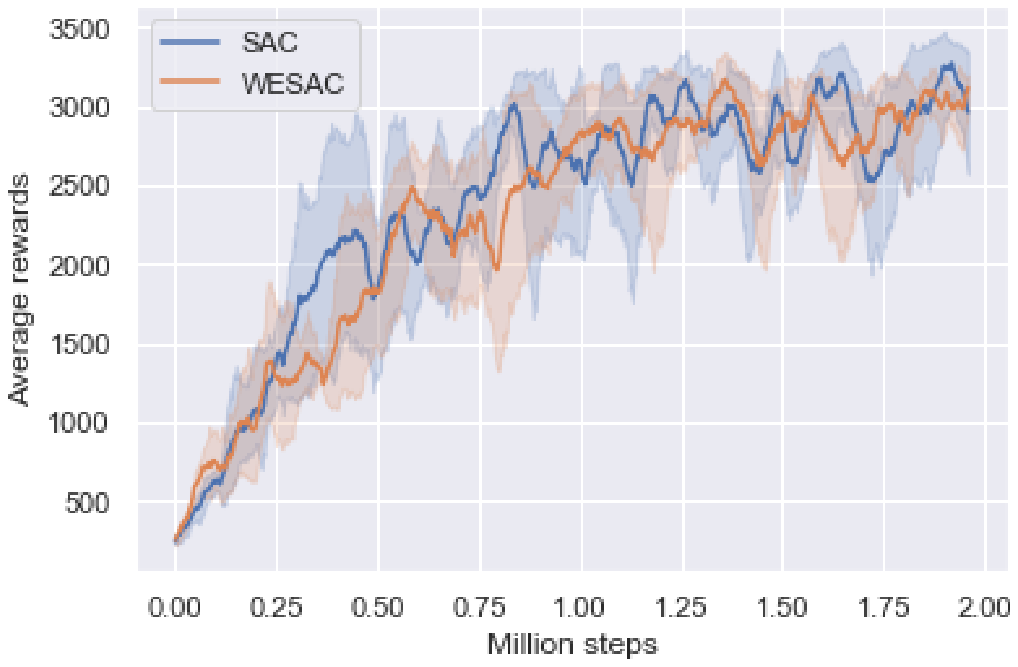}
\captionsetup{labelformat=empty}
\caption{(a) Hopper(v2)}
\end{minipage}
\begin{minipage}{0.499\textwidth}
\includegraphics[height=42mm, width=64mm]{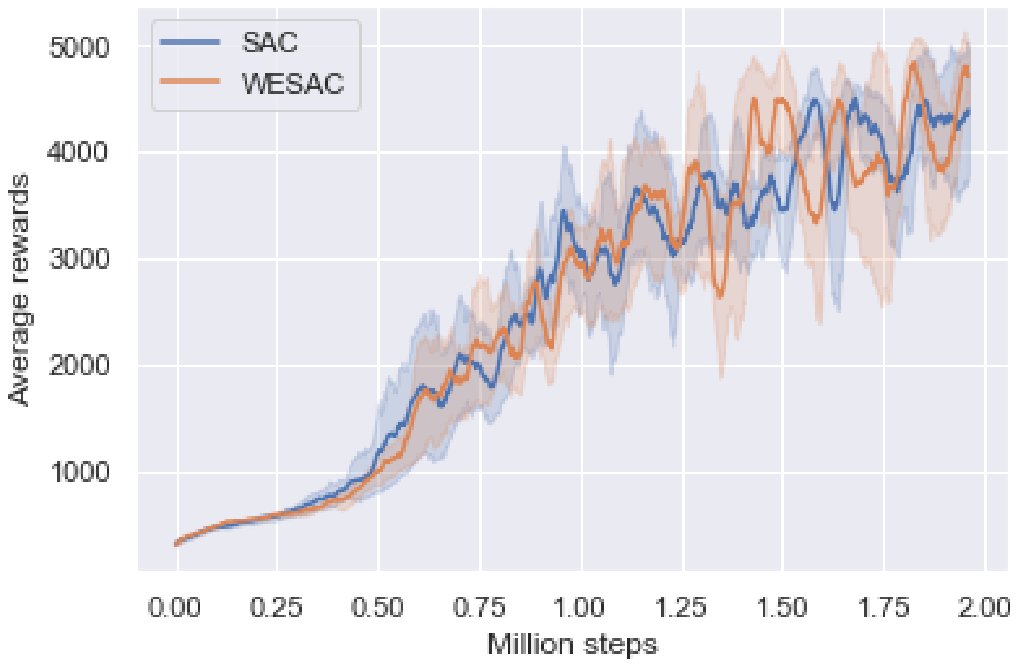}
\captionsetup{labelformat=empty}
\caption{(b) Humanoid(v2)}
\end{minipage}
%\end{figure}

%\begin{figure}[h!tbp]
\begin{minipage}{0.499\textwidth}
\includegraphics[height=42mm, width=64mm]{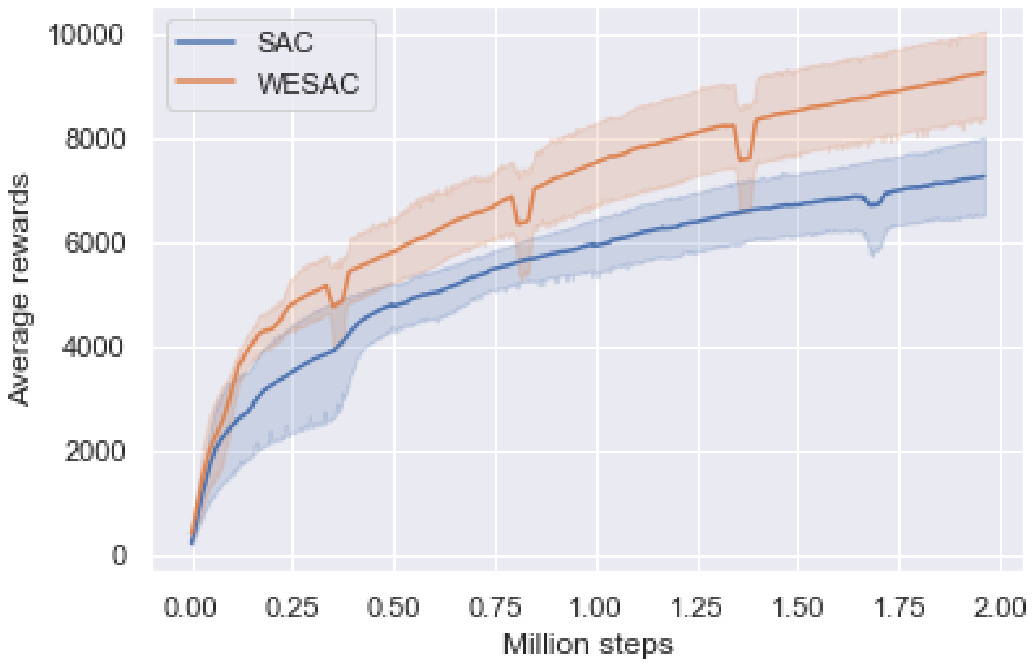}
\captionsetup{labelformat=empty}
\caption{(c) HalfCheeta(v2)}
\end{minipage}
\begin{minipage}{0.499\textwidth}
\includegraphics[height=42mm, width=64mm]{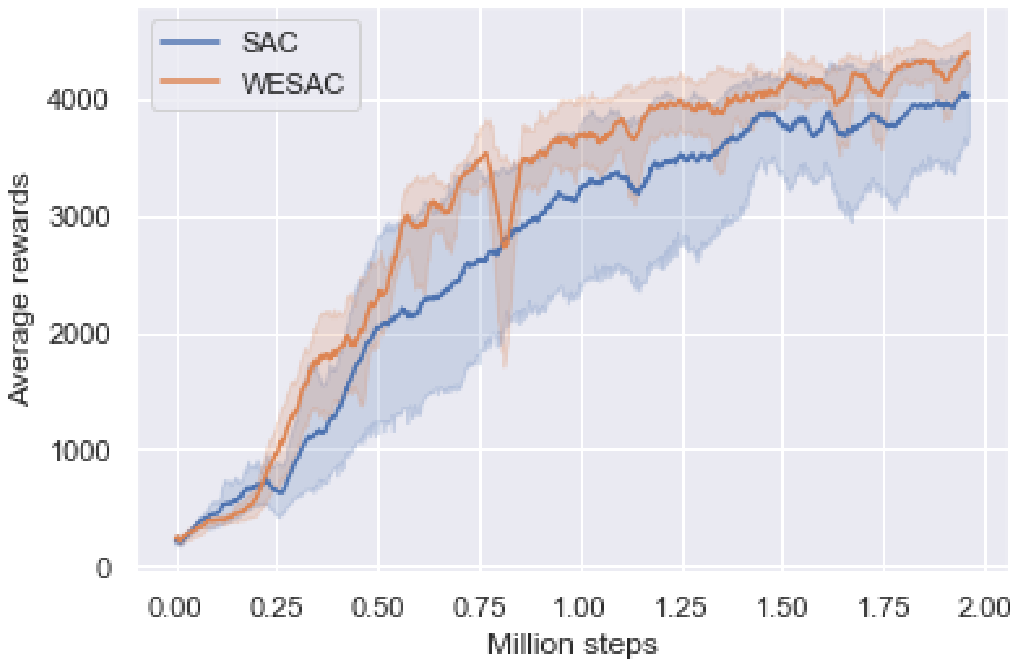}
\captionsetup{labelformat=empty}
\caption{(d) Walker2d(v2)}
\end{minipage}
%\end{figure}

%\begin{figure}[h!tbp]
\begin{minipage}{0.499\textwidth}
\includegraphics[height=42mm, width=64mm]{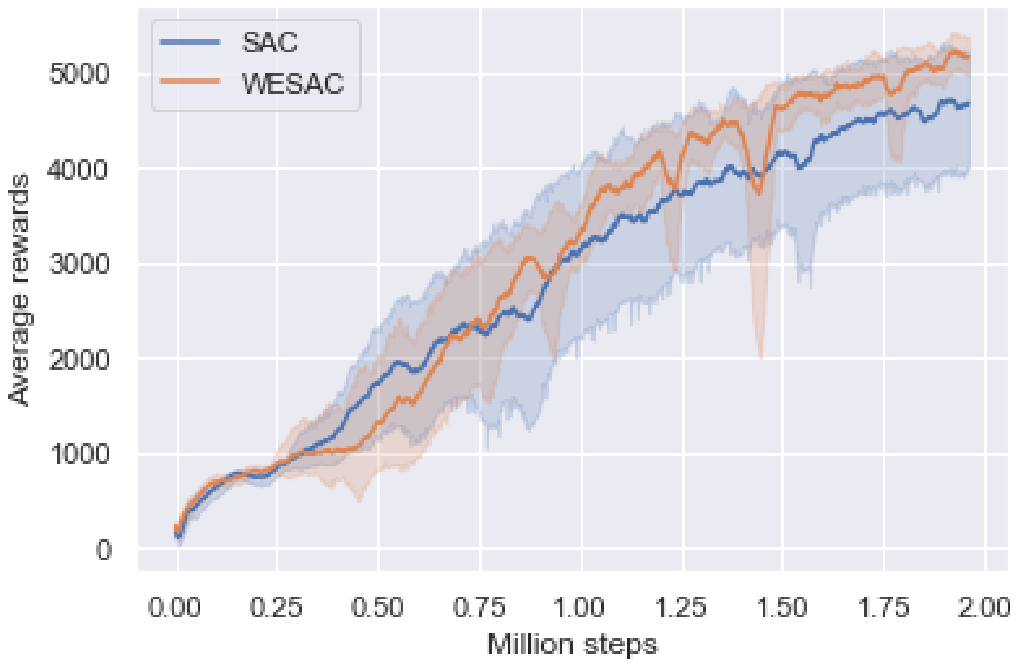}
\captionsetup{labelformat=empty}
\caption{(e) Ant(v2)}
\end{minipage}
\begin{minipage}{0.499\textwidth}
\includegraphics[height=42mm, width=64mm]{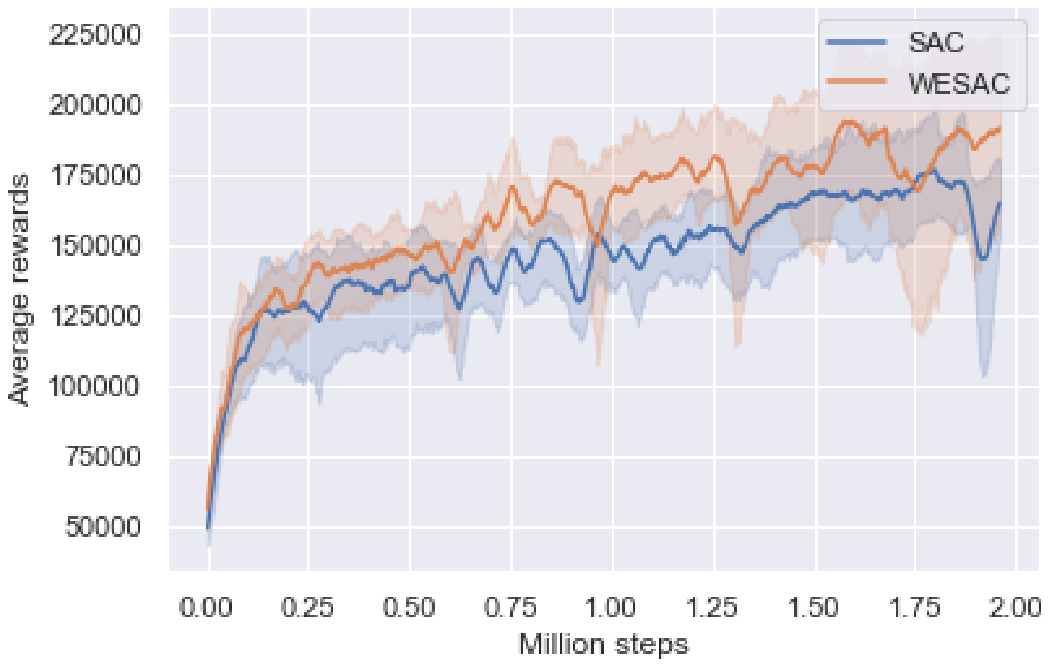}
\captionsetup{labelformat=empty}
\caption{(f) HumanoidStandup(v2)}
\end{minipage}
\captionsetup{labelformat=empty}
\caption{Figure 1: Each task is repeated for five times of different random seeds of two million steps for SAC and WESAC respectively. The solid lines in the figures show the moving-average-20 of the mean rewards of evaluation and the shadow areas indicate the variance. The reward scale of HumanoidStandup is $0.01$ to match the magnitude of rewards of the other tasks.}
\end{figure}

Figure 1 depicts the performance of WESAC compared with SAC. As shown in the figure, WESAC obtains similar results with SAC on the Hopper(v2) and Humanoid(v2). For the rest four tasks, WESAC outperforms SAC with respect to the average total rewards even though WESAC needs an extra step to calculate weight function \eqref{eq:weight}. 

One of the most fascinating properties of SAC is that it requires little hyper-parameter tuning. The mechanisms of the MuJoCo tasks are insufficiently studied in our experiment. In model-free reinforcement learning, our main goal is to obtain a more robust and stable algorithm rather than to demystify the mechanism of a certain environment. 

\section{Discussion and Conclusion}
In model-free reinforcement learning framework, the target to optimize varies according to different motivations. One of the commonly appreciated variations is to add a bonus term to rewards and the bonus $f$ is often related to exploration:
\begin{equation}
J(\pi)=\sum_{t=0}^{T} \mathbb{E}_{\left(s_{t}, a_{t}\right) \sim \rho_{\pi}}\left[\gamma^t(r\left(s_{t}, a_{t}\right)+\alpha\cdot f(s_t, a_t))\right]
\end{equation}
In this paper, we have extended idea f treating the bonus term as Shannon entropy to weighted entropy, which gives us more potential to link Q-function to auxiliary information. We proposed an idea here, which gives weights to both entropy and reward:
\begin{equation}
J^w(\pi)=\sum_{t=0}^{T} \mathbb{E}_{\left(s_{t}, a_{t}\right) \sim \rho_{\pi}}\left[\gamma^tw(s_t,a_t)(r\left(s_{t}, a_{t}\right)+\alpha \mathcal{H}\left(\pi\left(\cdot | s_{t}\right)\right))\right]
\end{equation}
Definitely, more careful studies are necessary to examine how to modify the objective rewards and update policies.

Guided by the mathematical derivation, we have developed an algorithm(WESAC) motivated by self-balancing intuition for exploration. Empirically we showed that WESAC's performance is no worse than SAC on canonical MuJoCo tasks in terms of average total rewards obtained in the first two million steps. More evaluations to judge whether our definition of the weight function encourages meaningful exploration will be presented in our future work.

\bibliography{main.bib}  % .bib
\newpage
\subsection*{Appendex A: axioms of Shannon entropy and weighted entropy}
Entropy, in physics and information theory, is the measure for randomness or dispersion. Usually, \textit{entropy} is regarded as Shannon entropy,
\begin{equation}\label{eq:entropy}
\mathcal{H}(X)=-\sum_{i=1}^{n} p_{i} \log _{2} p_{i}
\end{equation}
which is characterized by the so called Shannon-Khinchin axioms~\citep{khinchin2013mathematical}. Our notions follow~\citep{ilic2014generalized}.
Let the discrete distribution on the set of $n$ element be denoted with
\begin{equation*}
\Delta_{n} \equiv\left\{\left(p_{1}, \ldots, p_{n}\right) \bigg| p_{i} \geq 0, \sum_{i=1}^{n} p_{i}=1\right\}, \quad n>1
\end{equation*}
and let $\mathbb{R}^+$ denotes the set of positive real numbers. Shannon entropy of such distribution is
a function $\mathcal{H}_n(\Delta_n) \to \mathbb{R}^+$ following the properties as below:
\begin{itemize}
    \item $\mathcal{H}_n$ is continuous in $\Delta_n$;
    \item $\mathcal{H}_n$ takes its largest value for the uniform distribution, $U_{n}=(1 / n, \ldots, 1 / n), $i.e. $ \mathcal{H}_n(P) \leq \mathcal{H}_n(U_n)$, for any $P \in \Delta_n$;
    \item $\mathcal{H}_n$ is expandable $\mathcal{H}_{n+1}\left(p_{1}, p_{2}, \ldots, p_{n}, 0\right)=\mathcal{H}_{n}\left(p_{1}, p_{2}, \ldots, p_{n}\right)$ for all $\left(p_{1}, \dots, p_{n}\right) \in \Delta_{n}$
    \item $\mathcal{H}_n$ has the following composition rule: $P=\left(p_{1}, \ldots, p_{n}\right) \in \Delta_{n}, P Q=\left(r_{11}, r_{12}, \ldots, r_{n m}\right) \in \Delta_{n m}, n, m \in \mathbb{N}, n, m>1$ such that $p_{i}=\sum_{j=1}^{m} r_{i j}$ and $Q_{ | k}=\left(q_{1 | k}, \ldots, q_{m | k}\right) \in \Delta_{m}, q_{i | k}=r_{i k} / p_{k}$. Then, 
$\mathcal{H}_{n m}(P Q)=\mathcal{H}_{n}(P)+\mathcal{H}_{m}(Q | P)$,  where $
\mathcal{H}_m(Q | P)=\sum_{k} p_{k} \cdot \mathcal{H}_{m}\left(Q_{|k}\right)
$
\end{itemize}
Then the Shannon entropy is given by the definition \eqref{eq:entropy}. Roughly speaking, Shannon entropy is the function on the space of distribution and should be continuous and expandable, reaches maximum when it is a uniform distribution and follows a certain composition rule. 

For weighted entropy, since the probabilities of the events are attached by weights, the axiom with uniform distribution getting the maximum for Shannon entropy is discarded. Besides, since entropy is applied as a measure of uncertainty, and conditional probability or conditional entropy is not the main theme in maximum entropy reinforcement learning, the rule of composition for Shannon entropy can be set aside. Instead, when considering the weights along with the probabilities,i.e., the sequence of non-negative real-valued functions
$$
\left(\mathcal{H}_{n}\left(w_{1}, \ldots, w_{n} ; p_{1}, \dots, p_{n}\right)\right)_{1 \leq n<\infty}
$$
where every $\mathcal{H}_{n}\left(w_{1}, \ldots, w_{n} ; p_{1}, \ldots, p_{n}\right)$ is defined on the set $w_k \geq 0, p_k \geq 0, k = 1,2,...,n; \sum_{k=0}^np_k = 1$. Under the following axioms~\citep{guiacsu1971weighted},
\begin{itemize}
    \item $\mathcal{H}_{2}\left(w_{1}, w_{2} ; p, 1-p\right)$ is continuous w.r.t. $p$ on the interval $[0, 1]$;
    \item $\mathcal{H}_{n}\left(w_{1}, \ldots, w_{n} ; p_{1}, \ldots, p_{n}\right)$ is symmetric w.r.t. all pairs $(w_k, p_k), k = 1,2,...,n$;
    \item $\mathcal{H}_n$ is expandable in the following way:
    $
        I_{n+1}\left(w_{1}, \ldots, w_{n-1}, w^{\prime}, w^{\prime \prime} ; p_{1}, \ldots, p_{n-1}, p^{\prime}, p^{\prime \prime}\right) \\ 
        = I_{n}\left(w_{1}, \ldots, w_{n} ; p_{1}, \ldots, p_{n}\right)+p_{n} I_{2}\left(w^{\prime}, w^{\prime \prime} ; \frac{p^{\prime}}{p_{n}}, \frac{p^{\prime \prime}}{p_{n}}\right)
    $
    where $w_n = (p^\prime w^\prime+p^{\prime \prime}w^{\prime \prime})$, and $p_n = p^\prime + p^{\prime \prime}$;
    \item For the uniform distribution \(\mathcal{H}_{n}\left(w_{1}, \ldots, w_{n} ; \frac{1}{n}, \ldots,\frac{1}{n}\right)=L(n) \frac{w_{1}+\ldots+w_{n}}{n}\), where $L(n)$ is a positive number for every $n > 1$.
\end{itemize}
weighted entropy is defined as
\begin{align}
    \mathcal{H}^w(p) = \mathcal{H}_n(w_1,...,w_n; p_1,...,p_n) = -\sum_{k=1}^nw_kp_k\log p_k
\end{align}
where $w = (w_1,...,w_n),p=(p_1,...,p_n)$. 

\newpage
\subsection*{Appendix B: policy iteration}
Without loss of generality, we set $\alpha = 1$.

\textbf{Lemma 1 proof}:  Given policy $\pi_{\mathrm{old}}$, if
\begin{align}
    \mathrm{D}_{\mathrm{KL}}^w\left(\pi_{\mathrm{new}}\left(\cdot | s_{t}\right) \bigg\| \frac{\exp \left(\frac{Q^{\pi_{\mathrm{\mathrm{old}}}}\left(s_{t}, \cdot\right)}{ w(s_t,\cdot)}\right )}{Z^{\pi_{\mathrm{\mathrm{old}}}}\left(s_{t}\right)}\right) &\leq \mathrm{D}_{\mathrm{KL}}^w\left(\pi_{\mathrm{old}}\left(\cdot | s_{t}\right) \bigg\| \frac{\exp \left(\frac{Q^{\pi_{\mathrm{\mathrm{old}}}}\left(s_{t}, \cdot\right)}{ w(s_t,\cdot)}\right )}{Z^{\pi_{\mathrm{\mathrm{old}}}}\left(s_{t}\right) }\right)
\end{align}
\begin{align}\label{eq:constraint}
    \text{and }\int w(s_t,a_t)\pi_{\mathrm{new}}(a_t | s_t)da_t &=  \int w(s_t,a_t)\pi_{\mathrm{old}}( a_t | s_t)da_t
\end{align}
by the definition of weighted KL-divergence \eqref{eq:weighted-KL}, we get
\begin{align}\label{eq:inequality_V}
V^{\pi_{\mathrm{old}}}(s_t) &= \mathbb{E}_{a_t\sim\pi_{\mathrm{old}}}[Q^{\pi_{\mathrm{old}}}(s_t,a_t)- w(s_t,a_t)\log \pi_{\mathrm{old}}(a_t | s_t)] \nonumber  \\
&\leq \mathbb{E}_{a_t\sim\pi_{\mathrm{new}}}[Q^{\pi_{\mathrm{old}}}(a_t | s_t)- w(s_t,a_t)\log \pi_{\mathrm{new}}(a_t | s_t)]
\end{align}
And
\begin{align}\label{eq:inequality_Q}
&Q^{\pi_{\mathrm{old}}}(s_t, a_t) = r(s_t, a_t) +\gamma \mathbb{E}_{s_{t+1} \sim p}[V^{\pi_{\mathrm{old}}}(s_{t+1})] \nonumber \\
&\leq r(s_t, a_t) +\gamma E_{s_{t+1} \sim p}[\mathbb{E}_{a_{t+1}\sim\pi_{\mathrm{new}}}[Q^{\pi_{\mathrm{old}}}(s_{t+1}, a_{t+1})- w(s_{t+1},a_{t+1})\log \pi_{\mathrm{new}}(a_{t+1} | s_{t+1})]] \nonumber \\
&...\nonumber \\
&\leq Q^{\pi_{\mathrm{new}}}(s_t, a_t)
\end{align}
Therefore if the reward $r(\cdot,\cdot)$ and the weighed entropy for state $s_t$ are bound, we get the convergence of $Q$.

\textbf{Lemma 2 proof:} Updating rule \eqref{eq:renew} directly indicates inequality \eqref{eq:inequality_V}, therefore inequality \eqref{eq:inequality_Q} holds.

Repeated expansion of $Q^{\pi_\mathrm{old}}$ as above suggests the expected improvement of Q-function for the updated policy. In detail, the algorithm will sample $s_t$ from $\mathcal{D}$ and then $a_{t}$ from $\pi_\phi$. From \eqref{eq:policy}, the gradient of loss function $L_\phi(\pi)$ is
\begin{align}
    \nabla_\phi L_\phi(\pi) & =  \int_{a\in\mathcal{A}}\pi_\phi(a | s_t)\nabla_\phi\log \pi_\phi(a | s_t)\bigg[w(s_t,a)(\log \pi_\phi(a|s_t) + 1) - Q_{\theta}(s_t, a)\bigg]da
\end{align}
Compared to the loss defined by the weighted KL-divergence
\begin{align}
        L^\prime_\phi(\pi) = \mathrm{D}^w_{\mathrm{KL}}(\pi_\phi \big\|\exp(Q_\theta(s_t, \cdot)/w(s_t, \cdot)-\log Z_\theta))
\end{align}
The following deduction ignores the constraint \eqref{eq:constraint}:
\begin{align}
    \nabla_\phi L^\prime_\phi(\pi) & =  \int_{a\in\mathcal{A}}w(s_t,a)\pi_\phi(a|s_t)\nabla_\phi\log \pi_\phi(a|s_t)\bigg[\log \pi_\phi(a|s_t) - Q_{\theta}(s_t, a)\bigg]da
\end{align}
because $\int w\pi \nabla \pi = 0$ if the constraint $\int w \pi $ always equals a constant(which is infeasible in practice). In particular, when $w(s, a) \equiv 1$ for all $(s, a) \in \mathcal{S}\times\mathcal{A}$,  the gradient of $L^\prime_\phi(\pi)$ is the same as $L_\phi(\pi)$, which recovers the soft actor-critic algorithm.

\subsection*{Appendix C: hyperparameters}
We set $\eta = 0.01$, which is the smoothing coefficient for $\bar{\phi}$. For the reset of hyperparameters, see appendix D in~\citep{haarnoja2018soft}.

\end{document}